\documentclass[letterpaper, 10pt, journal, twoside]{IEEEtran}

\IEEEoverridecommandlockouts
\makeatletter

\let\proof\@undefined
\let\endproof\@undefined
\makeatother

\usepackage{amsmath}  
\usepackage{amssymb}  
\usepackage{amsthm}
\usepackage{amsfonts}
\usepackage{mathtools}

\usepackage{paralist}
\usepackage{fancyhdr}

\theoremstyle{definition}

\theoremstyle{remark}

\usepackage{verbatim}

\usepackage{multirow}
\usepackage{makecell}

\usepackage{color}
\usepackage{graphicx}

\usepackage{times}

\usepackage{algorithm}
\usepackage{etoolbox}\AtBeginEnvironment{algorithmic}{\scriptsize}
\usepackage[noend]{algpseudocode}

\usepackage[us]{datetime}
\usepackage{caption}
\usepackage[usenames,dvipsnames,table]{xcolor}

\definecolor{dark_green}{rgb}{0.0, 0.6, 0.0}

\usepackage{soul}
\newcommand{\rev}[1]{\sethlcolor{yellow!0}\hl{#1}} 

\usepackage{todonotes}
\usepackage{subfig}

\usepackage{latexsym}
\usepackage{color}

\usepackage[noadjust]{cite}
\usepackage{float}

\usepackage{tikz}
\usepackage{adjustbox}
\usepackage{pgfplots}
\pgfplotsset{compat=newest}
\usetikzlibrary{shapes,fit,arrows,calc,intersections,positioning,backgrounds,decorations.markings,patterns,external}

\tikzset{external/system call={latex \tikzexternalcheckshellescape -halt-on-error
    -interaction=batchmode -jobname "\image" "\texsource";
    dvips -o "\image".eps "\image".dvi;
ps2eps "\image.eps"}}

\let\labelindent\relax 
\usepackage[inline]{enumitem} 
\setenumerate[0]{label=\roman*.}

\usepackage{subfiles}
\usepackage{bm}

\tikzset{
  connect/.style args={(#1) to (#2) over (#3) by #4}{
    insert path={
      let \p1=($(#1)-(#3)$), \n1={veclen(\x1,\y1)},
      \n2={atan2(\x1,\y1)}, \n3={abs(#4)}, \n4={#4>0 ?180:-180}  in
      (#1) -- ($(#1)!\n1-\n3!(#3)$)
      arc (\n2:\n2+\n4:\n3) -- (#2)
    }
  },
}

\usepackage{scalerel}
\usetikzlibrary{svg.path}
\definecolor{orcidlogocol}{HTML}{A6CE39}
\tikzset{
  orcidlogo/.pic={
    \fill[orcidlogocol] svg{M256,128c0,70.7-57.3,128-128,128C57.3,256,0,198.7,0,128C0,57.3,57.3,0,128,0C198.7,0,256,57.3,256,128z};
    \fill[white] svg{M86.3,186.2H70.9V79.1h15.4v48.4V186.2z}
    svg{M108.9,79.1h41.6c39.6,0,57,28.3,57,53.6c0,27.5-21.5,53.6-56.8,53.6h-41.8V79.1z M124.3,172.4h24.5c34.9,0,42.9-26.5,42.9-39.7c0-21.5-13.7-39.7-43.7-39.7h-23.7V172.4z}
    svg{M88.7,56.8c0,5.5-4.5,10.1-10.1,10.1c-5.6,0-10.1-4.6-10.1-10.1c0-5.6,4.5-10.1,10.1-10.1C84.2,46.7,88.7,51.3,88.7,56.8z};
  }
}
\newcommand\orcidicon[1]{\href{https://orcid.org/#1}{\mbox{\scalerel*{
        \begin{tikzpicture}[yscale=-1,transform shape]
          \pic{orcidlogo};
        \end{tikzpicture}
}{|}}}}

\makeatletter
\let\NAT@parse\undefined
\makeatother

\usepackage{hyperref} 
\hypersetup{
  colorlinks,
  citecolor=black,
  filecolor=black,
  linkcolor=black,
  urlcolor=black,
  pdfauthor={},
  pdfsubject={},
  pdftitle={}
}

\setlist[itemize]{noitemsep, nosep}

\usepackage{siunitx}
\usepackage{ifthen}

\usepackage{booktabs}
\renewcommand{\arraystretch}{0.99}
\renewcommand{\baselinestretch}{0.99}
\usepackage{tablefootnote}
\makeatletter
\def  \input@path{{./../fig/},{./fig/}}
\makeatother

\title{%
  Large-Scale Exploration of Cave Environments by Unmanned Aerial Vehicles
}

\author{
  Pavel Petr\'{a}\v{c}ek$_a^{\orcidicon{0000-0002-0887-9430}}$,
  V\'{i}t Kr\'{a}tk\'{y}$_a^{\orcidicon{0000-0002-1914-742X}}$,
  Mat\v{e}j Petrl\'{i}k$_a^{\orcidicon{0000-0002-5337-9558}}$,
  Tom\'{a}\v{s} B\'{a}\v{c}a$_a^{\orcidicon{0000-0001-9649-8277}}$, 
  Radim Kratochv\'{i}l$_b^{\orcidicon{0000-0003-1884-4052}}$, and
  Martin Saska$_a^{\orcidicon{0000-0001-7106-3816}}$%
\thanks{%
  Manuscript received March 1, 2021; Revised May 6, 2021; Accepted July 8, 2021.
  This paper was recommended for publication by Editor Pauline Pounds upon evaluation of the Associate Editor and Reviewers’ comments.}
  \thanks{%
    The work was supported
    by the Czech Science Foundation (GA\v{C}R) under research project no. 20-29531S,
    by CTU grant no. SGS20/174/OHK3/3T/13,
    by the Defense Advanced Research Projects Agency (DARPA), and
    by OP VVV funded project CZ.02.1.01/0.0/0.0/16 019/0000765 "Research Center for Informatics".}
  \thanks{%
  $_a$ Department of Cybernetics, Faculty of Electrical Engineering, Czech Technical University in Prague, 166 36 Prague 6, Czech Republic {\tt\footnotesize\{\href{mailto:pavel.petracek@fel.cvut.cz}{pavel.petracek}|\href{mailto:vit.kratky@fel.cvut.cz}{vit.kratky}|\href{mailto:matej.petrlik@fel.cvut.cz}{matej.petrlik}|\href{mailto:tomas.baca@fel.cvut.cz}{tomas.baca}|\newline\href{mailto:martin.saska@fel.cvut.cz}{martin.saska}\}@fel.cvut.cz},\newline
  $_b$ Brno University of Technology, Faculty of Civil Engineering, Institute of Geodesy, Czech Republic {\tt\footnotesize\href{mailto:r_kratochvil@fce.vutbr.cz}{r\_kratochvil@fce.vutbr.cz}}.}
  \thanks{Digital Object Identifier (DOI): see top of this page.}
}

\begin{document}

\newcommand{\PREPRINTYEAR}{2021}
\newcommand{\PREPRINTPUBLISHER}{IEEE Robotics and Automation Letters}
\newcommand{\DOI}{10.1109/LRA.2021.3098304}

\markboth{\PREPRINTPUBLISHER, \PREPRINTYEAR. DOI: \DOI}{}

\fancyhead{}
\chead{©\PREPRINTPUBLISHER, \PREPRINTYEAR. DOI: \DOI}
\pagestyle{fancy}
\thispagestyle{plain}

\onecolumn
\pagenumbering{gobble}
{
  \topskip0pt
  \vspace*{\fill}
  \centering
  \LARGE{%
    © \PREPRINTYEAR~\PREPRINTPUBLISHER\\\vspace{1cm}
    Personal use of this material is permitted.
    Permission from \PREPRINTPUBLISHER~must be obtained for all other uses, in any current or future media, including reprinting or republishing this material for advertising or promotional purposes, creating new collective works, for resale or redistribution to servers or lists, or reuse of any copyrighted component of this work in other works.\\\vspace*{1cm}DOI: \DOI}
    \vspace*{\fill}

}
\twocolumn 
\pagenumbering{arabic}

\maketitle

\begin{abstract}
  This paper presents a self-contained system for the robust utilization of aerial robots in the autonomous exploration of cave environments to help human explorers, first responders, and speleologists.
  The proposed system is generally applicable to an arbitrary exploration task within an unknown and unstructured subterranean environment and interconnects crucial robotic subsystems to provide full autonomy of the robots.
  Such subsystems primarily include mapping, path and trajectory planning, localization, control, and decision making.
  Due to the diversity, complexity, and structural uncertainty of natural cave environments, the proposed system allows for the possible use of any arbitrary exploration strategy for a single robot, as well as for a cooperating team.
  A multi-robot cooperation strategy that maximizes the limited flight time of each aerial robot is proposed for exploration and search~\&~rescue scenarios where the homing of all deployed robots back to an initial location is not required.
  The entire system is validated in a comprehensive experimental analysis comprising of hours of flight time in a real-world cave environment, as well as by hundreds of hours within a state-of-the-art virtual testbed that was developed for the DARPA Subterranean Challenge robotic competition.
  Among others, experimental results include multiple real-world exploration flights traveling over \SI{470}{\meter} on a single battery in a demanding unknown cave environment.
\end{abstract}

\begin{IEEEkeywords}
  Aerial Systems: Applications; Field Robots; Aerial Systems: Perception and Autonomy; Multi-Robot Systems; Mapping
\end{IEEEkeywords}

\vspace{-0.5em}
\section*{Multimedia Materials}
\label{sec:multimedia_materials}
The paper is supported by the multimedia materials available at \href{http://mrs.felk.cvut.cz/papers/ral-2021-caves}{mrs.felk.cvut.cz/papers/ral-2021-caves}.
\rev{%
  The implementation is also publicly available at \mbox{\href{https://github.com/ctu-mrs}{github.com/ctu-mrs}}.}



\section{Introduction}

\IEEEPARstart{H}{uman} exploration of complex cave systems has occurred for thousands of years.
However, there are still entire cave systems and individual subterranean voids, shafts, and cavities that are yet uncovered.
This is primarily due to the dangerous nature of subterranean exploration in environments like natural caves, although man-made cellars, drainages, and mines pose similar risks.
These environments contain sediments such as debris, rocks, sand, clay, ice, decomposed organic matter, human waste, and even various forms of speleothems in limestone caves.
Considering the absolute darkness, lack of GNSS signals, flowing and dripping water, humid air, and the possible presence of poisonous gases, wind gusts, hanging ropes, and wildlife, there is excessive risk to the lives of human explorers in the exploration of new environments, as well as in search~\&~rescue missions.
Given the current state-of-the-art technology in robotics, many dangerous areas of subterranean systems are safely reachable using mobile robots, with the greatest focus being on vertical exploration using aerial vehicles.
In contrast to human exploration, the use of such technology presents several advantages in the form of accessibility, safety, speed, instantaneous environment visualization, and precise quantification.
On the other hand, challenges to the operation of mobile robots in such an environment lies in the uncertainty, lack of light, high humidity, and diversity of space in the form of narrow and/or low passages, canyons, large domes, high chimneys, and deep abysses.

The challenges to deployment of aerial vehicles in subterranean environments with respect to robot control, communication, sensor fusion, and positioning are described thoroughly in~\cite{8877114}.
These specific challenges continue to be relevant even after substantial progress in the field of mobile robotics.
However, in contrast to~\cite{8877114}, our motivation includes minimizing the need for communications required for operator control and instead focuses on the full autonomy of robots and autonomous cooperation among members of a robotic team.
The restriction of communication in subterranean environments introduces challenges to the maximization of system robustness and the use of efficient decision making in the form of adaptable exploration strategies in harsh unknown environments.
\vfill


\begin{figure}[t]
  \centering
    \includegraphics[width=1.00\columnwidth]{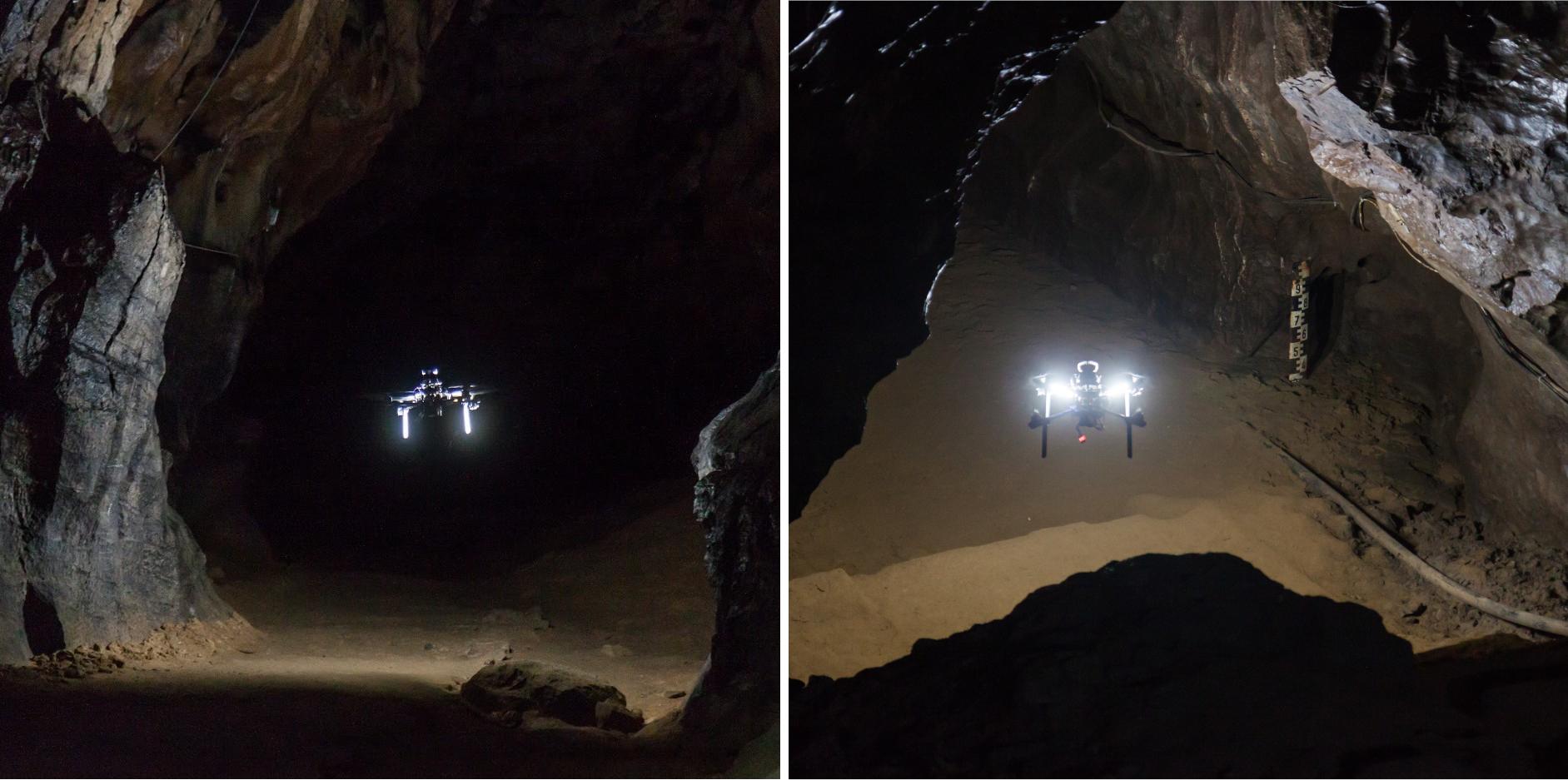}
  \caption{Robotic exploration of the Bull Rock Cave (central Moravian Karst, Czech Republic) by a fully autonomous aerial vehicle.}
  \label{fig:byci_skala_motivation}
  \vspace{-0.5em}
\end{figure}


\vspace{-0.2em}
\subsection{Related work}

In a non-robotic context, wild caves are explored by modernly termed \textit{cavers}.
However, the human surveying and mapping of caves is known to have existed for thousands of years for purposes ranging from dwelling to speleology.
The significance of cave exploration and cave mapping to scientific research is a thoroughly studied inquiry in literature, e.g., in~\cite{caves_non_robotic}.

In the work presented here, we focus mainly on the robotic point of view within the scope of the application domain.
\rev{%
  One of the first cave-mapping approaches using robotic solutions was proposed in~\mbox{\cite{Zlot2014THREEDIMENSIONALMM}}, where the authors employed hand-held laser scanners, which are limited in speed, accuracy, and safety.
In the context of mobile robotics, topics like the automatic control of an unstable dynamic system such as an aerial multi-rotor vehicle~{\cite{NASCIMENTO2019129}}, the fusion of inertial, visual, and laser information for localization and mapping~{\cite{s20072068}}, and path planning in dynamic environments~{\cite{ZHAO201854}} have been addressed in order to achieve faster and safer methodology than mapping done with hand-held devices, as proposed in~{\cite{Zlot2014THREEDIMENSIONALMM}}.}

Within the scope of subterranean environments, the DARPA Subterranean Challenge competition has pushed the state of the art of autonomous exploration in human-made mines~\cite{9197082,8981594,darpa_upenn,petrlik2020robust}.
\rev{%
  Although these systems have provided interesting solutions with great potential, the authors of~\mbox{\cite{9197082,8981594,darpa_upenn,petrlik2020robust}} rely on the predictable structure of underground mines, such as using the protraction of human-made tunnels to mark the furthest depth data as frontiers or predefining turns at junctions in~{\cite{darpa_upenn}}}.
Since the complexity and diversity of natural caves is extensive, more robust solutions with a minimum number of environmental assumptions are required.
This was tackled in~\cite{7991499} where the authors introduced a possible way for applying autonomous drones as a technology to assist speleologists and archaeologists.
Although an interesting read, the proposed methods only constitute a preliminary discussion that presents neither novel technology nor applied results.
A similar discussion focusing on the state of robotic problems within the application of subterranean exploration with UAVs is presented in~\cite{10.1117/12.2266316}.
In contrast to~\cite{7991499}, the authors of~\cite{10.1117/12.2266316} present a set of preliminary experiments in laboratory conditions and two dimensional space.
Unfortunately, the assumption of a planar world is highly restrictive within the scope of real-world deployment due to the complex character of natural subterranean environments.

The precise localization of mobile robots is crucial to autonomous navigation in such complex environments.
Among existing state-of-the-art literature, the LOCUS algorithm~\cite{palieri2020locus} achieves the lowest localization error at the cost of high computational demands.
Unlike with ground robots, this method might be unsuitable for aerial robots as the computational resources on lightweight UAVs are scarce due to their limited payload.
In~\cite{funabiki2020range}, the authors demonstrated that localization performance can be further improved by dropping range beacons.
This is a viable strategy for heterogeneous robotic teams, but unfeasible for teams of only lightweight UAVs.

\rev{%
  The use of robotic teams for cooperative exploration has been addressed mostly in planar worlds with recurrent connectivity constraints~{\cite{5509803}} or with the requirement of a centralized element~{\cite{1435481}}.
 A similarly defined task to our problem of team homing --- respecting intermittent communication, need for decentralization, and limited operation time of aerial robots --- is proposed in~{\cite{7139494}}, where the robots gather and share data during the mission and return all the way back to the base before their operation times out.
  In contrast to~{\cite{7139494}}, we propose homing coordination that lands each aerial robot at a position expanding a communication relay graph, thereby increasing the time for mere exploration in tasks where return to the starting position is not required.
  Related to the scope of search~\&~rescue, the authors in~{\cite{10.1145/2750675.2750683}} propose to re-position robots in a relay-chain formation to enable data transmission over longer distances once an object of interest is found.
  Our solution reports the position of the objects once the explorer robot connects to the relay graph during homing.}
The recently developed fast exploration technique in~\cite{zhou2021fuel} maximizes explored volume over battery-limited flight time.
The method is based on data only from an RGBD camera with a limited field of view (FoV).
In comparison to LiDAR-based methods, we have experimentally verified that RGBD cameras are sub-optimal sensors for the exploration of large-scale caves due to their limited range and~FoV.


\vspace{-0.2em}
\subsection{Contributions}

First, we propose a fully autonomous system enabling multi-modal mapping, fast and efficient planning with sensoric field-of-view constraints for safe movement in 3D, robust localization, and adaptable decision making.
Second, a multi-robot cooperation for the efficient homing of a team of autonomous explorer robots is proposed.
Third, the system has been validated through hundreds of hours of testing in a state-of-the-art virtual testbed developed for the DARPA Subterranean Challenge robotic competition, as well as through hours of flight time in the real world.
To the best of our knowledge, the presented large-scale experimental deployment of autonomous aerial robots in a natural cave environment goes beyond the current state of the art in autonomous robotics.
Lastly, we present and share the experience obtained during this comprehensive experimental deployment that was carried out in close cooperation with speleologists.



\section{Experience Gained}
\label{sec:experience_gained}

\subsection{Speleology motivation}

From the speleological point of view, aerial systems are crucial for pushing exploratory state-of-the-art methods to provide assistance in efficient scouting of difficult-to-access areas in vertical environments, as well as for the quick inspection of known areas using onboard sensors only.
These systems minimize risks for humans by reducing the need to climb or to swim in cold water reservoirs, and also through the detection of poisonous gases or even radioactive waste.
Furthermore, this enables the preservation and protection of natural environments against human influence, including ancient sediment forms, floor dripstone formations, paleontological and archaeological sites, and sources of potable water.

In contrast to well-established methods of subterranean documentation (i.e., theodolite and level/distance meter, compass, and clinometer), modern technology employs stationary and mobile laser scanners to produce a dense 3D model of the environment.
Due to the complexity of natural environments, the use of stationary scanners is time-consuming because of the necessity of eliminating occluded spaces.
Although hand-held mobile scanners are more time-efficient in this context, their use is limited to areas accessible to humans.
This limitation opens the door for mobile robotics which is able to tackle this challenge and to provide optimized 3D mapping.
State-of-the-art mapping in such environments reaches decimeter level precision, which is less precise than stationary scanners, yet sufficient for the majority of speleological needs.
Moreover, the common issue of mapping drift accumulation in long-corridor spaces can be minimized using reference measurements by precise stationary scanners or man-measured control points to obtain accurate results.

\vspace{-0.2em}
\subsection{System requirements}

The primary prerequisite of a team of aerial explorers that can be deployed in caves involves the ability to adapt to diverse, unknown environments lacking sources of light and access to GNSS.
This general description requires the abilities to
\begin{itemize}
  \item be deployed in constrained cavities, as well as in open caverns of natural caves,
  \item map and visualize the environment in a fast, quantified manner in the form of dense point clouds and image streams,
  \item seamlessly infuse an arbitrary exploration strategy for more efficient mission operation within the scope of individual environments (policy selection is discussed in~\autoref{sec:exploration}),
  \item return to the mission operator and promptly \rev{visualize} the environment for human supervision, and
  \item maximize operation capabilities in terms of coverage when a team of robots is employed.
\end{itemize}


\vspace{-0.2em}
\subsection{Depth estimation in high humidity}

The performance of the PMD pico flexx time-of-flight (ToF) camera and the Intel Realsense D435 stereo camera have been analyzed as complementary sensors to the primary LiDAR for the purpose of improving the sensory FoV coverage.
Although ToF cameras generally outperform stereo cameras in terms of distance measurement precision and density of measurement points~\cite{pece2011three}, the high humidity typically present in natural caves causes dispersion of light emitted from ToF cameras by small water droplets.
This effect significantly degrades the acquired measurements.
As was verified empirically, ToF cameras can produce false-negative measurements of obstacles situated behind clouds of water droplets.
The use of stereo cameras (e.g., Realsense) is recommended for its robustness to environmental conditions within natural caves.
Nevertheless for large cave systems, such a sensor needs to be combined with 3D LiDARs in order to comply with the requirements of speleologists and first responders.




\section{System Architecture}
\label{sec:system_architecture}

\begin{figure}[t]
  \centering
  \vspace{-0.7em}
\pgfdeclarelayer{background}
\pgfdeclarelayer{foreground}
\pgfsetlayers{background,main,foreground}

\tikzstyle{block}=[draw, rounded corners, text centered, minimum height=2.0em, fill=white, fill opacity=1.0, text opacity=1.0]
\def\nodedst{2.0cm}

\begin{tikzpicture}[auto, node distance=3.0cm, >=latex, font=\scriptsize]

  \renewcommand{\arraystretch}{0.6}


  \begin{pgfonlayer}{foreground}
    \node [block] (mission) {\begin{tabular}{c}
        \footnotesize Mission \\
        \footnotesize control
    \end{tabular}};
    \node [block, right of=mission, node distance=2.1cm] (navigation) {
        \begin{tabular}{c}
        \footnotesize Navigation \&\\
        \footnotesize planning
        \end{tabular}};
    \node [block, right of=navigation, shift = {(0.0, -0.0)}, node distance=2.4cm] (tracker) {
        \begin{tabular}{c}
          \footnotesize Reference\\
          \footnotesize tracker
      \end{tabular}};
    \node [block, right of=tracker, node distance=2.2cm] (control) {
        \begin{tabular}{c}
          \footnotesize Reference\\
          \footnotesize controller
      \end{tabular}};
    \node [block, above of=control, fill=white, node distance=1.25cm, shift={(-0.15,0)}] (attitude) {
        \begin{tabular}{c}
          \footnotesize Attitude rate\\
          \footnotesize controller
        \end{tabular}
      };
    \node [block, left of=attitude, node distance=2.05cm] (actuator) {\footnotesize Actuators};
    \node [block, below of=control, node distance=\nodedst, shift = {(-0.15, -0.0)}] (state) {\footnotesize State estimation};
    \node [block, below of=navigation, shift = {(0.5, -0.0)}, node distance=0.97cm] (mapping) {\footnotesize Mapping};
    \node [block, left of=state, node distance=3.0cm] (loam) {\footnotesize LOAM};
    \node [block, below of=mission, node distance=\nodedst] (lidar) {\footnotesize LiDAR};
    \node [block, above of=navigation, node distance=1.25cm] (imu) {\footnotesize IMU};
    \node [block, left of=mapping, node distance=1.5cm] (rgbd) {\footnotesize RGBD};
  \end{pgfonlayer}


  \draw [->] (mission.east) -- node[above, shift={(0, 0)}] {\footnotesize} (navigation.west);
  \draw [->] (navigation.east) -- node[above, shift={(0.00, 0)}] {\footnotesize $T_{d}$} node[below, shift={(0.05, 0)}] {} (tracker.west);

  \draw [->] (tracker.east) -- node[above, shift={(0, 0)}] {\footnotesize $\mathbf{x}_d$} (control.west);
  \draw [->] (control.113) -- node[right, shift={(0, 0)}] {\footnotesize $\mathbf{\omega}_d$} (attitude.south);
  \draw [->] (attitude.west)+(0.0,0) -- (actuator.east);


  \node[inner sep=0,minimum size=0,left of=lidar, shift={(2.0,-0.55)}] (k) {}; 
  \draw [-] (imu.west) node[below, shift={(-1.25, 0.0)}] {\footnotesize$\mathbf{\omega}_{\text{IMU}}, \mathbf{R}_{\text{IMU}}$} -| (k);
  \draw [->] (k) -| (state.south);

  \draw [->] (mapping.north) -- ($(navigation.south)+(0.5,0.0)$);
  \draw [->] (lidar.east) -- node[above, shift={(0, 0)}] {} (loam.west);
  \draw [->] (lidar.east) + (2,0) -| node[above, shift={(0, 0)}] {} (mapping.south);
  \draw [->] (loam.east) -- node[above, shift={(0, 0)}] {\footnotesize $\mathbf{R}_{\text{LOAM}}$} node[below, shift={(0, 0)}] {\footnotesize $\mathbf{r}_{\text{LOAM}}$} (state.west);
  \draw [->] (loam.north) |- (mapping.east);
  \draw [->] (rgbd.east) -- (mapping.west);

  \draw [->] (state.67) -- node[left, shift={(0.0, -0.1)}] {\footnotesize $\mathbf{r}$, $\mathbf{\omega}$, $\mathbf{R}$} (control.south);
  \draw [->] (state.67) +(0.0,0.30) -| (tracker.south)+(0.0,-0.60) -- (tracker.south);

  \begin{pgfonlayer}{background}
    \path (mission.west |- mission.north)+(-0.10,0.4) node (a) {};
    \path (navigation.south -| navigation.east)+(+0.05,-0.1) node (b) {};
    \path[fill=blue!10,rounded corners, draw=black!70, densely dotted]
    (a) rectangle (b);
  \end{pgfonlayer}
  \node [rectangle, above of=mission, node distance=1.7em, shift={(0.40,-0.05)}] (text_control) {\footnotesize \textbf{High-level planning}};

  \begin{pgfonlayer}{background}
    \path (tracker.west |- tracker.north)+(-0.05,0.4) node (a) {};
    \path (control.south -| control.east)+(+0.1,-0.1) node (b) {};
    \path[fill=red!10,rounded corners, draw=black!70, densely dotted]
    (a) rectangle (b);
  \end{pgfonlayer}
  \node [rectangle, above of=tracker, node distance=1.7em, shift={(0.31,-0.05)}] (text_control) {\footnotesize \textbf{Tracking \& control}};

  \begin{pgfonlayer}{background}
    \path (mission.west |- mapping.north)+(-0.10,0.1) node (a) {};
    \path (state.south -| state.east)+(+0.1,-0.1) node (b) {};
    \path[fill=green!10,rounded corners, draw=black!70, densely dotted]
    (a) rectangle (b);
  \end{pgfonlayer}
  \node [rectangle, above of=lidar, node distance=1.7em, shift={(-0.15,0.45)}] (text_control) {\footnotesize \textbf{Perception}};

  \begin{pgfonlayer}{background}
    \path (imu.west |- attitude.north)+(-2.6,0.15) node (a) {};
    \path (imu.south -| attitude.east)+(+0.1,-0.10) node (b) {};
    \path[fill=orange!10,rounded corners, draw=black!70, densely dotted]
    (a) rectangle (b);
  \end{pgfonlayer}
  \node [rectangle, above of=imu, shift={(-2.3,-0.3)}, node distance=1.7em] (autopilot) {\footnotesize \textbf{Autopilot}};


\end{tikzpicture}
  \caption{
    Individual interconnected modules form the system architecture of the autonomous explorer robot.
    The \textit{High-level planning} modules focus on achieving the mission objectives by generating references for the \textit{Tracking \& control} modules based on the map built by the \textit{Perception} modules. This also provides a state estimate for closing the control feedback loop.
    All modules except the \textit{Autopilot} group are handled by the main onboard computer.
    \label{fig:system_architecture}
  }
  \vspace{-0.7em}
\end{figure}
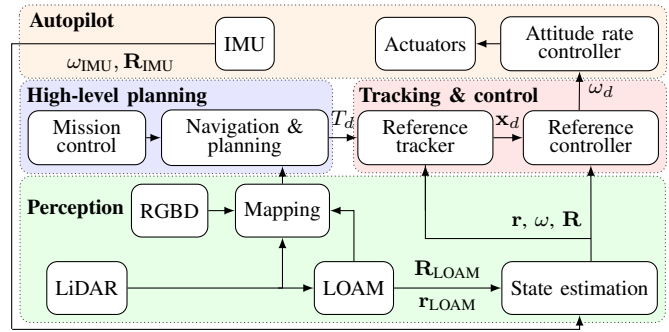

The system of the proposed autonomous explorer robot is divided into multiple groups of individual interconnected modules to be described in this section.
All components and their relations are visualized in \autoref{fig:system_architecture}.

\vspace{-0.2em}
\subsection{Perception}
The perception of the proposed system is based on a multi-channel \textit{LiDAR} sensor that is used for both building the spatial representation of the surrounding environment in the \textit{Mapping} module, as well as for the motion estimation in the \textit{LOAM} module.
Obtaining the full-state estimate is realized within the \textit{State estimation} module, where multiple sources of incomplete state measurements are fused together using a bank-of-filters~estimator.

The vertical navigation capabilities of the system can be greatly improved by equipping the robot with vertically-facing \textit{RGBD} cameras that are able to fill in the blind spots in the limited vertical FoV of the \textit{LiDAR}.
Apart from navigation, these optional sensors may be used for detecting objects of interest in caves in search~\&~rescue scenarios or for visual documentation of newly explored cave systems.

\subsubsection{LiDAR}
Even though our system is not tied to a specific LiDAR model, there are certain important parameters that can affect the performance and capabilities of the platform.


To reliably stabilize the UAV, the time delay of the estimated state must stay below the threshold of a certain critical value depending on the type of controller and gains.
When this threshold is exceeded, the UAV begins oscillating and eventually automatically lands when the control error is too large to continue the mission safely.
We have found experimentally that for most combinations of localization methods and controllers, the critical value ranges from \SIrange{100}{200}{\milli\second}.
Thus, \SI{10}{\hertz} is the lowest rotation frequency that can be used without employing methods of delay compensation.


The typical values of a vertical field of view (VFoV) of 3D LiDARs are in the \SIrange{30}{90}{\degree} range.
The higher VFoV values improve vertical mobility in constrained spaces, however with a low VFoV, it is impossible to safely navigate narrow vertical shafts as it is not known whether the space above the UAV is free and safe to fly through, or whether it contains an obstacle.

\subsubsection{RGBD}
The regions above and below the UAV that are not covered by the LiDAR can be captured using a depth camera or by spinning the LiDAR sensor around a vector that is orthogonal to the axis of scanning, as seen in~\cite{csiro_darpa}.
However, such a solution adds additional weight to the sensor, which decreases the available flight time.
A blind spot also still remains as part of the laser rays is blocked by the frame of the UAV.
Alternatively, lightweight depth cameras can be mounted on opposite sides of the body frame in order to cover most blind spots of the LiDAR.
Additional sensing modality is gained by combining an RGB and depth camera in a single sensor (RGBD) with a slight weight increase.



\subsubsection{Localization}
\label{sec:localization}
For localization of the UAV, we have adapted the LOAM algorithm~\cite{loam}.
This state-of-the-art method is very precise (\SI{0.55}{\percent} translation error \cite{Geiger2012CVPR}) while attaining real-time performance.
In our adapted version of the open-source implementation, the algorithm is optimized on CPU and employs parallel computing, which enables us to deploy and use the localization in the real-time position control feedback loop onboard fast-moving aerial vehicles.

\subsubsection{Mapping}
\label{sec:mapping}
The LOAM-based algorithm builds a sparse internal representation of the environment consisting of edge and planar features.
However, this sparse map is unsuitable for navigation purposes. 
Additionally, the LOAM map does not consider the probabilistic nature of the sensor, nor does it distinguish free and unknown space.
Both of these factors are necessary in exploration techniques for reliable navigation and consistent frontier selection.

In the proposed system, the environment is represented by a dense probabilistic volumetric map, which consists of cubic cells with one of 3 states: free, occupied, or unknown.
The map is kept in the octree structure to facilitate the Bayesian integration of new measurements and efficient access to individual cells of the probabilistic map.
The high-level systems, such as grid-based path planning or inter-robot map registration, also benefit from quick access to the dense environment representation.
This approach is capable of multi-modal fusion by integrating the data from all available onboard sensors and outputting point-cloud measurements.
If high-level path planning is constrained by the field of view of onboard sensors (tackled in~\cite{8392775} and also in~\autoref{sec:path_planning}), the multi-modality of mapping enables arbitrary movement in 3D.

\subsubsection{Sensor processing}

The targeted subterranean environments may have high humidity or may contain large clouds of whirling dust.
The water and dust particles can then produce erroneous measurements for the LiDAR-based sensors.
Assuming a partial reflection from water or dust particles and a large energy dissipation of distant reflections, these erroneous measurements can be filtered with respect to the measured intensity of returning light rays.
As has been empirically verified, a simple threshold-based filtration over the intensity channel within the local neighborhood of the sensor is sufficient for filtering out false-positive measurements.
The idea of the local filtration is to filter out particles gusting through the surrounding air due to the aerodynamic influence of the propellers.
Although the cutoff threshold of the intensity magnitude is environment-specific, filtering out measurements below the 10$^{th}$ percentile of the intensity distribution per each laser scan proved to be a reliable solution, even in the dustiest real-world environments.
Such processing is unavailable for camera-based systems that may require thorough, computationally-expensive solutions to overcome these~challenges.


\subsubsection{\mbox{State estimation}}
The reference controller (see \autoref{sec:reference_controller}) requires a position estimate of the UAV body frame in the world frame $\mathbf{r}=\left(x, y, z\right)$, the velocity of the body frame $\dot{\mathbf{r}}$, rotation $\mathbf{R}$ from the UAV body frame to the world frame, and angular velocity $\bm{\omega}$ in the body frame in order to close the feedback loop.
\rev{The LOAM localization method provides 6-DoF pose estimate, i.e., $\mathbf{r}_{\text{LOAM}}$, $\mathbf{R}_{\text{LOAM}}$, which are fused in the \textit{State estimation} block with interoceptive measurements from the \textit{IMU} of the \textit{Autopilot} to obtain the rest of the state variables.}

The details about the estimation process are described in~\cite{baca2020mrs}.
Nevertheless, it is worth highlighting the importance of the fusion of orientation $\mathbf{R}_{\text{IMU}}$ and $\mathbf{R}_{\text{LOAM}}$ in cave environments.
While $\mathbf{R}_{\text{IMU}}$ is very precise and without delay, the heading of the UAV (i.e., the measured direction of the body-fixed, forward-facing axis) is unreliable due to the presence of ferromagnetic ores in the cave rocks that cause deviations in the magnetometer measurements.
\rev{By correcting these errors with the heading from $\mathbf{R}_{\text{LOAM}}$ in the estimation process, the resulting orientation $\mathbf{R}$ is robust to changes in the erratic magnetic field in subterranean environments.}

\vspace{-0.2em}
\subsection{Tracking \& control}
\label{sec:tracking_control}
The safe navigation of constrained environments with low obstacle clearance imposes the requirements of precise trajectory tracking with minimal control error, as any deviation from the desired state could potentially result in a collision.
The \textit{Reference controller} is responsible for minimizing the control error around the desired control reference that is provided by the \textit{Reference tracker}.
The controller outputs an attitude rate reference for the low-level \textit{Attitude rate controller} in the \textit{Autopilot}.

\subsubsection{Reference tracker}
The \textit{Reference tracker} is essential in providing the \textit{Reference controller} with smooth and feasible references to ensure a safe flight.
The tracker based on the model predictive control (MPC) simulates an ideal virtual model of the UAV with constrained translational states up to jerk, together with heading and heading rate.
The input can be either a single pair of desired 3D position $\mathbf{p}_d$ and heading $\eta_d$, or a trajectory $T_d$ in the form of a sequence of such pairs with a specified sampling rate. 
The full state of the virtual model is then sampled at \SI{100}{\hertz}, and $\mathbf{r}_d$, $\dot{\mathbf{r}}_d$, $\ddot{\mathbf{r}}_d$, $\dot{\ddot{\mathbf{r}}}_d$, $\eta_d$, $\dot{\eta_d}$ are passed to the \textit{Reference controller} as reference $\mathbf{x}_d$.

\subsubsection{Reference controller}
\label{sec:reference_controller}
The agile \textit{SE(3)} geometric state feedback controller \cite{lee2010geometric} minimizes the position and velocity errors.
To compensate imperfect calibration and external forces acting upon the UAV, the controller is extended with the body and world disturbance terms described in~\cite{baca2020mrs}.
The output attitude rate reference $\bm{\omega}_d$ is tracked by the \textit{Autopilot}.

\vspace{-0.2em}
\subsection{Path planning}
\label{sec:path_planning}


The planning approach used to safely navigate through apriori unknown environments must fulfill requirements of real-time responsiveness and efficient global planning in order to fully exploit the limited flight time of UAVs.
\rev{For this purpose, fast iterative post-processing is applied to the output of an optimal grid-based planner in order to increase the UAV-obstacle distance above a minimum threshold~{\cite{kratky2021exploration}.}}
The grid-based planner and the iterative post-processing do not apply an optimistic assumption that the unknown space is collision-free.
Although this visibility-constrained precondition requires high sensory coverage around the robot to allow for arbitrary movement in 3D, it consequently prevents collisions of the trajectory being followed, even if replanning would fail.
This methodology improves safety and robustness of the overall flight, allows for deployment in completely unknown environments without any apriori information, and permits seamless navigation in open spaces, as well as safe movement through narrow passages.

Common grid-based planning methods require pre-processing of an employed map representation, such as determining and applying the 3D distance transform for obstacle growing.
This may introduce significant computational overhead by bottle-necking system performance, as the map must then be processed in every planning step.
Such a computationally expensive task contradicts the requirements for responsiveness within evolving dynamic environments.
To minimize the overall time required for a single planning iteration, a local KD-tree representation of the environment is used to decide the feasibility of particular cells within a voxel grid.
This approach shifts the largest load from the pre-processing phase to the planning phase, which is beneficial especially to shorter plans that require searching only a small part of the environment.
The low computational demands of the applied planning approach enable frequent replanning the global plan, which is also crucial for the efficient use of newly-discovered collision-free space.

To effectively exploit the limited flight time of aerial explorers, all mid-flight stops are eliminated by computing in parallel the next exploration goal during path following.
The path to the next goal is efficiently appended to the rest of the current reference trajectory $T_d$ using the prediction horizon of the MPC (see~\autoref{sec:tracking_control}).
The need for precise locomotion control in complex natural caves makes uniform path-sampling unfeasible with respect to the dynamic constraints of a UAV and fast, collision-free trajectory tracking.
Therefore, the reference trajectory $T_d$ provided by the \textit{Navigation \& planning} module to the \textit{Reference tracker} is computed based on the following process. 

Given the dynamical constraints of the robot, the generated path is uniformly sampled with a sampling distance adapted to the maximum velocity magnitude $v_{max}$ of the UAV.
Based on this initial trajectory $T_i$, the required acceleration magnitudes $a_n$ between consequent transition points are computed by velocity differentiation as
\begin{equation}\label{eq:sampling_acc}
  a_n(k) = \frac{||\mathbf{v}_i(k+1) - \mathbf{v}_i(k)||_2}{t_s},
\end{equation}
where $\mathbf{v}_i(k)$ is the required velocity vector for transition from a transition point $t_i(k)$ to $t_i(k+1)$ on the initial trajectory $T_i$ and $t_s$ is a constant sampling period.
The new velocity for a $k$-th segment is then given by
\begin{equation}\label{eq:sampling_vel} 
  v_k = \left\{\begin{array}{lll}
      \text{max}\left(v_{max}\frac{a_{max}}{a_n}, v_{min}\right) & \text{if} & a_n(k) > a_{max},\\
      v_{max} & \text{if} & a_n(k) \leq a_{max},
    \end{array}\right.
\end{equation} 
where the minimum velocity $v_{min}$ serves as a parameter balancing the precision and the time needed for trajectory tracking.
By this step, the velocities for particular segments are set so that the maximum velocity is applied in straight segments, while lower velocities are applied in curved segments of any given path.

To further improve trajectory sampling and to achieve smoother changes in velocities, the sampling distance on particular segments is computed so that the motion along each segment has the constant acceleration
\begin{equation}
  \overline{a}_k = \frac{|v_{k+1} - v_k|}{t_{acc, k}},
\end{equation} 
where $t_{acc,k}$ is the time available for acceleration on the \mbox{$k$-th} segment.
The time $t_{acc,k}$ is obtained from the length $l_k$ of the segment $k$ and the required change of the velocity.
The number of transition points $N_k$ on the $k$-th segment of the initial trajectory $T_i$ is given as
\begin{equation}
\def\arraystretch{1.8}
  N_k = \left\{\begin{array}{cll}
      \left \lceil \frac{l_k}{v_k t_s} \right \rceil & \text{if} & \overline{a}_k = 0, \\
      \left \lceil \frac{t_{acc}}{t_s} \right \rceil & \text{if} & \overline{a}_k > 0,
  \end{array}\right.
\def\arraystretch{1.0}
\end{equation}
where the desired constant acceleration is adapted to meet the velocity $v_{k+1}$ at the end of each segment as
\begin{equation}
  a_{k} = \frac{\overline{a}_k}{N_k t_s}.
\end{equation}
The sequence of sampling distances for the $k$-th segment of $T_i$ is then given by 
\begin{equation}\label{eq:sampling_dists}
  d_{k,i} = v_k t_s + i a_{k} {t_s}^2,\,\, i \in \{1, \cdots, N_k\}.
\end{equation}

\rev{The trajectory sampled with sampling distances defined by{~\eqref{eq:sampling_dists}} is passed to the \textit{Reference tracker}~{\cite{baca2020mrs}} as a reference trajectory $T_d$ in order to generate a feasible reference $\mathbf{x}_d$ for the \textit{Reference controller}.}
Despite its simplicity, the described sampling method achieves better results within the scope of the proposed application than the optimization-based trajectory generation methods proposed in~\cite{richter2016polynomial, burri2015real-time}.
In contrast to the proposed method, the problem in~\cite{richter2016polynomial, burri2015real-time} is defined in such a way that the exact positions of all the path waypoints must be visited, generating significantly slower trajectories.






\section{Exploration Policy}
\label{sec:exploration}

Cave environments are naturally diverse and require various different mission strategies suitable for specific environments.
Deriving the optimal policy is thereby dependent on various factors, such as the expected mission output, mission-specific constraints, the complexity and the specifics of the environment, and the number of available robots.
For this reason, our system is designed so that any arbitrary policy can be utilized within the scope of an autonomous mission.

Nevertheless, two exploratory mission types are of the most use in practice: deep cave exploration and full-coverage exploration.
These missions are used for scouting previously uncovered areas in order to obtain a general overview of the environment, monitor environmental changes such as gas leaks, detect natural water reservoirs, discover new possible passages, or assess the structural state of cavern walls and other objects of interest.
The former approach maximizes the explored volume of space in the entire environment, while the latter minimizes the blind spots missed by onboard cameras with a constrained FoV. 


The capabilities of a robotic mission are furthered with the use of multiple cooperating robots.
To \rev{show} an example of such improvement using a team of agents as opposed to a single agent, a homing strategy that maximizes the flight time of aerial robots during a multi-robotic mission is proposed in the following subsection.
\rev{%
During the proposed coordination, continuous exploration is not assumed and distance-constrained ad-hoc communication is used.
The robots are homogeneous and generate their behaviors in a decentralized manner based on their current state and the available information from other robots (only positions in a shared frame are required).}

\vspace{-0.2em}
\subsection{Multi-robot homing strategy}

A cooperative operation maximizing the flight time of a multi-robot team is proposed for applications where homing all the deployed robots to an initial location is not required.
This strategy is suitable for tasks where the possible gained information is superior to the cost of the robots, such as in search~\&~rescue scenarios.
This method assumes there is access to a low-bandwidth communication link among any two robots within an omnidirectional communication radius.

To maximize the flight time, the robots utilize local communication to plan the homing path such that a group of robots is able to build up a communication tree with the base station as the root communication node.
This allows the robots to optimize their flight time by navigating back to a location in the proximity of another communication node (a landed robot, base station, or self-sustaining communication node deployed by other robots) when the battery capacity becomes drained.
This entire homing strategy is showcased in an example scenario for two independent robots in \autoref{fig:graph_cave}.

In the proposed strategy, each robot constructs a navigation homing tree using nodes created from the set of past poses of the robot.
This online-built tree has edges valued by the required flight time between two nodes and is used to estimate required homing time to the proximity of a communication node. 
The pose nodes are connected such that each path leaf-to-communication is the shortest (see~\autoref{fig:graph_cave}a).
A homing path is constructed recursively as a sequence of tree nodes from the current robot position (a leaf) to the nearest communication node, with the landing position being within communication range of the nearest communication node (see~\autoref{fig:graph_cave}b).
The tree is shared among the robots deployed in the same mission.
The knowledge from the previous explorers is integrated to prolong their flight time (see~\autoref{fig:graph_cave}c), thus causally maximizing the time capacity for the exploration task.
When a communication node (e.g., a robot landing pose) is integrated into the homing tree, it is linked exclusively to another communication node to join the retranslation chain (see~\autoref{fig:graph_cave}d).
Consequently, the parents of neighboring pose nodes are updated so that each pose node has a parent with the minimal accumulated cost to any communication node (see~\autoref{fig:graph_cave}b and \autoref{fig:graph_cave}d).
The process of inserting pose nodes as well as communication nodes into the homing tree is described in Alg.~\ref{alg:homing_tree}.

\begin{figure}[t]
  \centering
  \definecolor{color_gen_b}{rgb}{0.22, 0.2, 0.502}
\definecolor{color_gen_a}{rgb}{0.737,0.165,0}
\definecolor{color_comm}{rgb}{0, .522, .243}

\begin{tikzpicture}[font=\small]

  \tikzstyle{node_comm} = [circle, draw, scale=0.45, color=color_comm, fill=color_comm!30, text=black, thick]
  \tikzstyle{node_comm_range} = [circle, draw, dashed, scale=7.0, color=black, fill=color_comm, draw opacity=0.2, fill opacity=0.05]
  \tikzstyle{node_gen_a} = [draw=color_gen_a, fill=color_gen_a!30, cross out, very thick, scale=0.45]
  \tikzstyle{node_gen_b} = [node_gen_a, draw=color_gen_b, fill=color_gen_b!30]
  \tikzstyle{node_gen_a_op} = [node_gen_a, opacity=0.55]

  \tikzstyle{edge_comm} = [draw=color_comm, fill=color_comm!30, thin]
  \tikzstyle{edge_gen_a} = [draw=color_gen_a, fill=color_gen_a!30, thin]
  \tikzstyle{edge_gen_b} = [edge_gen_a, draw=color_gen_b, fill=color_gen_b!30]
  \tikzstyle{edge_gen_a_op} = [edge_gen_a, opacity=0.3]

  \begin{axis}[
      name=ax1,
      width=0.30\textwidth,
      enlarge x limits=0.0,
      enlarge y limits=0.0,
      hide axis
    ]

    \coordinate (label) at (80, -82);
    \coordinate (basestation) at (-85, -85);
    \coordinate (comm_a) at (-40, -50);
    \coordinate (comm_b) at (-59, 10);
    \coordinate (a1) at (-50, -70);
    \coordinate (b1) at (-32, -35);
    \coordinate (c1) at (-12, 5);
    \coordinate (d1) at (32, -1);
    \coordinate (e1) at (60, -30);
    \coordinate (f1) at (15, 38);
    \coordinate (g1) at (48, 70);
    \coordinate (h1) at (90, 80);
    \coordinate (i1) at (110, 80);
    \coordinate (a2) at (-40, -70);
    \coordinate (b2) at (-37, -28);
    \coordinate (c2) at (-45, 9);
    \coordinate (d2) at (-71, 35);
    \coordinate (e2) at (-97, 60);
    \coordinate (f2) at (-110, 60);

    \addplot[on layer=axis background]
    graphics[xmin=-100,ymin=-100,xmax=100,ymax=100] {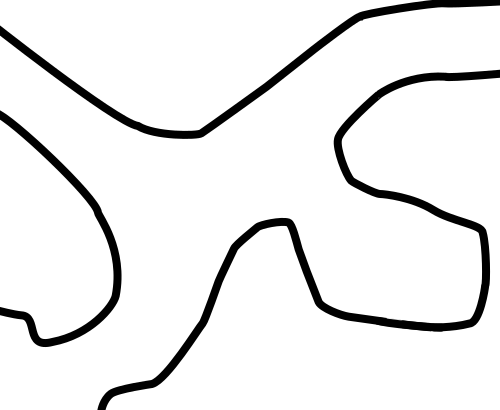};

    \node[node_comm] (n_basestation) at (basestation) {\textbf{B}};
    \node[node_comm_range] at (basestation) {};

    \draw[edge_gen_a] (n_basestation.20) -- (a1);
    \draw[edge_gen_a] (a1) -- (b1);
    \draw[edge_gen_a] (b1) -- (c1);
    \draw[edge_gen_a] (c1) -- (d1);
    \draw[edge_gen_a] (c1) -- (f1);
    \draw[edge_gen_a] (d1) -- (e1);
    \draw[edge_gen_a] (f1) -- (g1);
    \draw[edge_gen_a] (g1) -- (h1);
    \draw[edge_gen_a] (h1) -- (i1);

    \node[node_gen_a] at (a1) {};
    \node[node_gen_a] at (b1) {};
    \node[node_gen_a] at (c1) {};
    \node[node_gen_a] at (d1) {};
    \node[node_gen_a] at (e1) {};
    \node[node_gen_a] at (f1) {};
    \node[node_gen_a] at (g1) {};
    \node[node_gen_a] at (h1) {};

    \node[scale=0.8] at (label) {\textbf{a)}};

    \draw [black, ultra thick] (100,100) -- (100,-100);
    \draw [black, ultra thick] (-100,-100) -- (100,-100);

  \end{axis}

  \begin{axis}[
      name=ax2,
      width=0.30\textwidth,
      at=(ax1.east), anchor=west,
      enlarge x limits=0.0,
      enlarge y limits=0.0,
      hide axis
    ]

    \addplot[on layer=axis background]
    graphics[xmin=-100,ymin=-100,xmax=100,ymax=100] {fig/graph_cave_bg.png};

    \node[node_comm] (n_basestation) at (basestation) {\textbf{B}};
    \node[node_comm_range] at (basestation) {};
    
    \draw[edge_gen_a] (comm_a.-100) -- (a1);
    \draw[edge_gen_a] (comm_a.45) -- (b1);
    \draw[edge_comm] (comm_a.200) -- (n_basestation.30);
    \draw[edge_gen_a] (b1) -- (c1);
    \draw[edge_gen_a] (c1) -- (d1);
    \draw[edge_gen_a] (c1) -- (f1);
    \draw[edge_gen_a] (d1) -- (e1);
    \draw[edge_gen_a] (f1) -- (g1);
    \draw[edge_gen_a] (g1) -- (h1);
    \draw[edge_gen_a] (h1) -- (i1);

    \node[node_comm, scale=0.9] at (comm_a) {$\mathbf{C_1}$};
    \node[node_comm_range] at (comm_a) {};

    \node[node_gen_a] at (a1) {};
    \node[node_gen_a] at (b1) {};
    \node[node_gen_a] at (c1) {};
    \node[node_gen_a] at (d1) {};
    \node[node_gen_a] at (e1) {};
    \node[node_gen_a] at (f1) {};
    \node[node_gen_a] at (g1) {};
    \node[node_gen_a] at (h1) {};
    
    \node[scale=0.8] at (label) {\textbf{b)}};

    \draw [black, ultra thick] (-100,-100) -- (100,-100);

  \end{axis}

  \begin{axis}[
      name=ax3,
      width=0.30\textwidth,
      at=(ax1.south), anchor=north,
      enlarge x limits=0.0,
      enlarge y limits=0.0,
      hide axis
    ]

    \addplot[on layer=axis background]
    graphics[xmin=-100,ymin=-100,xmax=100,ymax=100] {fig/graph_cave_bg.png};

    \node[node_comm] (n_basestation) at (basestation) {\textbf{B}};
    \node[node_comm_range] at (basestation) {};

    \draw[edge_gen_a_op] (comm_a.-100) -- (a1);
    \draw[edge_gen_a_op] (comm_a.45) -- (b1);
    \draw[edge_comm] (comm_a.200) -- (n_basestation.30);
    \draw[edge_gen_a_op] (b1) -- (c1);
    \draw[edge_gen_a_op] (c1) -- (d1);
    \draw[edge_gen_a_op] (c1) -- (f1);
    \draw[edge_gen_a_op] (d1) -- (e1);
    \draw[edge_gen_a_op] (f1) -- (g1);
    \draw[edge_gen_a_op] (g1) -- (h1);
    \draw[edge_gen_a_op] (h1) -- (i1);
    \draw[edge_gen_b] (comm_a.90) -- (a2);
    \draw[edge_gen_b] (b2) -- (comm_a.90);
    \draw[edge_gen_b] (comm_a.100) -- (c2);
    \draw[edge_gen_b] (c2) -- (d2);
    \draw[edge_gen_b] (d2) -- (e2);
    \draw[edge_gen_b] (e2) -- (f2);

    \node[node_comm, scale=0.9] at (comm_a) {$\mathbf{C_1}$};
    \node[node_comm_range] at (comm_a) {};

    \node[node_gen_a_op] at (a1) {};
    \node[node_gen_a_op] at (b1) {};
    \node[node_gen_a_op] at (c1) {};
    \node[node_gen_a_op] at (d1) {};
    \node[node_gen_a_op] at (e1) {};
    \node[node_gen_a_op] at (f1) {};
    \node[node_gen_a_op] at (g1) {};
    \node[node_gen_a_op] at (h1) {};
    \node[node_gen_b] at (a2) {};
    \node[node_gen_b] at (b2) {};
    \node[node_gen_b] at (c2) {};
    \node[node_gen_b] at (d2) {};
    \node[node_gen_b] at (e2) {};

    \node[scale=0.8] at (label) {\textbf{c)}};

    \draw [black, ultra thick] (100,100) -- (100,-100);

  \end{axis}

  \begin{axis}[
      name=ax4,
      width=0.30\textwidth,
      at=(ax2.south), anchor=north,
      enlarge x limits=0.0,
      enlarge y limits=0.0,
      hide axis
    ]

    \addplot[on layer=axis background]
    graphics[xmin=-100,ymin=-100,xmax=100,ymax=100] {fig/graph_cave_bg.png};

    \node[node_comm] (n_basestation) at (basestation) {\textbf{B}};
    \node[node_comm_range] at (basestation) {};

    \draw[edge_gen_a_op] (comm_a.-100) -- (a1);
    \draw[edge_gen_a_op] (comm_a.45) -- (b1);
    \draw[edge_comm] (comm_a.200) -- (n_basestation.30);
    \draw[edge_gen_a_op] (b1) -- (c1);
    \draw[edge_gen_a_op] (c1) -- (d1);
    \draw[edge_gen_a_op] (c1) -- (f1);
    \draw[edge_gen_a_op] (d1) -- (e1);
    \draw[edge_gen_a_op] (f1) -- (g1);
    \draw[edge_gen_a_op] (g1) -- (h1);
    \draw[edge_gen_a_op] (h1) -- (i1);
    \draw[edge_gen_b] (comm_a.90) -- (a2);
    \draw[edge_gen_b] (b2) -- (comm_a.90);
    \draw[edge_gen_b] (d2) -- (e2);
    \draw[edge_gen_b] (e2) -- (f2);

    \draw[edge_gen_b] (comm_b.0) -- (c2);
    \draw[edge_gen_b] (comm_b.90) -- (d2);
    \draw[edge_comm] (comm_b.-90) -- (comm_a.100);

    \node[node_comm_range] at (comm_a) {};
    \node[node_comm_range] at (comm_b) {};
    \node[node_comm, scale=0.9] at (comm_a) {$\mathbf{C_1}$};
    \node[node_comm, scale=0.9] at (comm_b) {$\mathbf{C_2}$};

    \node[node_gen_a_op] at (a1) {};
    \node[node_gen_a_op] at (b1) {};
    \node[node_gen_a_op] at (c1) {};
    \node[node_gen_a_op] at (d1) {};
    \node[node_gen_a_op] at (e1) {};
    \node[node_gen_a_op] at (f1) {};
    \node[node_gen_a_op] at (g1) {};
    \node[node_gen_a_op] at (h1) {};
    \node[node_gen_b] at (a2) {};
    \node[node_gen_b] at (b2) {};
    \node[node_gen_b] at (c2) {};
    \node[node_gen_b] at (d2) {};
    \node[node_gen_b] at (e2) {};

    \node[scale=0.8] at (label) {\textbf{d)}};

  \end{axis}
 
  \begin{axis}[
      name=leg,
      at=(ax1.north east), anchor=south,
      width=0.60\textwidth,
      height=0.11\textwidth,
      enlarge x limits=0.0,
      enlarge y limits=0.0,
      font=\footnotesize,
      hide axis
    ]

    \addplot [opacity=0.0] coordinates {(0,0)(1,1)}; 
    \coordinate (marker_bs) at (0.05, 0.5);
    \node[node_comm] (a) at (marker_bs) {\textbf{B}};
    \node[right of=a, anchor=west, node distance=0.08cm] (b) {base station};
    \node[node_comm, right of=b, node distance=2.3cm] (a) {$\mathbf{C}$};
    \node[right of=a, anchor=west, node distance=0.10cm] (b) {communication node (range incl.)};
    \node[node_gen_a, right of=b, black, node distance=5.1cm] (a1) {};
    \node[node_gen_a, right of=a1, black, node distance=0.7cm] (a2) {};
    \draw[edge_gen_a, black] (a1) -- (a2);
    \node[right of=a2, anchor=west, node distance=0.05cm] (b) {homing tree};

  \end{axis}

\end{tikzpicture}
  \vspace{-1em}
  \caption{
    An example scenario of the homing strategy for two robots (red and blue) that maximizes flight time by landing at feasible positions while building a communication chain to a base station. 
  }
  \label{fig:graph_cave}
  \vspace{-2.0em}
\end{figure}
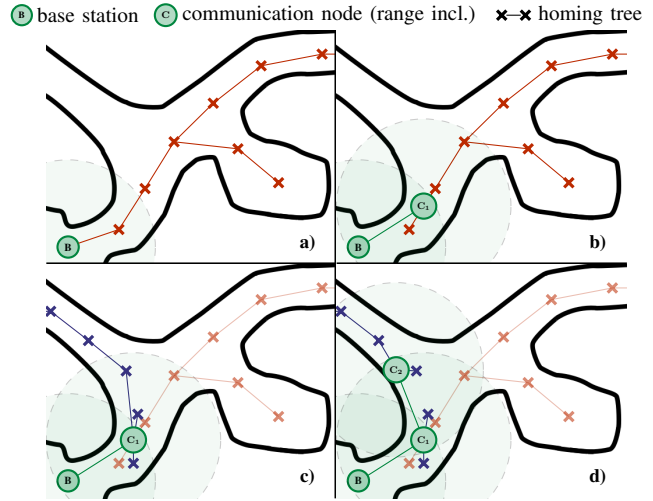

\begin{figure}[t]
  \centering
  \footnotesize
  \begin{minipage}[!t]{1.0\columnwidth}
    \begin{algorithm}[H]

      \algdef{SE}[SUBALG]{Indent}{EndIndent}{}{\algorithmicend\ }%
      \algtext*{Indent}
      \algtext*{EndIndent}

      \algnewcommand\algorithmicinput{\textbf{Input:~}}
      \algnewcommand\Input{\State\algorithmicinput}%

      \captionsetup{labelformat=empty}
      \caption{\footnotesize{\textbf{Algorithm 1:}
      Insertion of a node into the onboard-built homing tree.
      Function cost($n_a, n_b$) returns an estimate of flight time among nodes $n_a$ and $n_b$,
      function accumulatedCost($n_a$) returns the required flight time from node $n_a$ to the nearest communication node, and
      function freeRay($n_a, n_b$) returns true if a linear path between nodes $n_a$ and $n_b$ is collision-free in 3D.
      }}
      \label{alg:homing_tree}

      \begin{algorithmic}[1]

        \Procedure{InsertNodeToHomingTree}{}

        \Input

        \Indent

        \State $\mathrm{N}$\Comment{Node to be inserted}
        \State $\mathcal{C}, \mathcal{P}$\Comment{Sets of communication and pose nodes}
        \State $d_e$\Comment{Minimum edge length}

        \EndIndent
        
        \If{$\mathrm{N}.type$ == COMMUNICATION}

        \State $\mathrm{N}.parent \gets \arg\min_{c \in \mathcal{C}} \text{cost}(\mathrm{N}, c)$

        \For{$p \in \mathcal{P}$}\Comment{Update parents of neighboring pose nodes}
        
          \If{$\text{cost}(\mathrm{N}, p) < \text{accumulatedCost}(p)$}
          \State $p.parent \gets \mathrm{N}$
          \EndIf

        \EndFor
          
        \State $\mathcal{C} \gets \mathcal{C} \cup \mathrm{N}$

        \Else

        \State $\mathcal{V} \gets \mathcal{C} \cup \mathcal{P}$

        \If {$\min_{v \in \mathcal{V}} \left(||\mathrm{N}-v||_2\right) \geq d_e$}

        \State $\overline{\mathcal{V}} = \left\{v \mid \text{freeRay}(\mathrm{N},v), \forall v \in \mathcal{V} \right\}$

        \If{$\overline{\mathcal{V}} \not= \emptyset$}
        
        \State $\mathrm{N}.parent \gets \arg\min_{v \in \overline{\mathcal{V}}}\;[\text{cost}(\mathrm{N},v) + \text{accumulatedCost}(v)]$
        \State $\mathcal{P} \gets \mathcal{P} \cup \mathrm{N}$

        \EndIf

        \EndIf

        \EndIf



        \EndProcedure

      \end{algorithmic}
    \end{algorithm}
  \end{minipage}
\vspace{-2.0em}
\end{figure}
\section{Experimental Analysis}
\label{sec:experimental_analysis}
The entire proposed system has been validated through hours of flight time in the real world, as well as in hundreds of hours in various virtual subterranean environments.
The results of these experimental analyses are presented hereafter.

\vspace{-0.2em}
\subsection{Real-world environment}
\label{sec:real_world_environment_exp}

\begin{figure}[htb]
  \centering
  \input{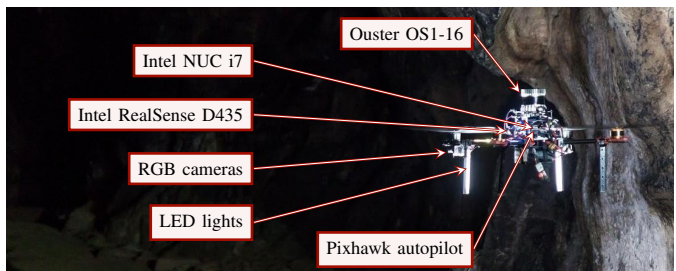}
  \vspace{-1.6em}
  \caption{
    General hardware components of an autonomous explorer robot.
    All data are processed and reasoned over with an onboard processing unit.
    The main source of data comes from the top-mounted LiDAR.
  }
  \label{fig:byci_skala_hw}
  \vspace{-1.6em}
\end{figure}

To analyze the properties of the system, a fully autonomous aerial robot (see~\autoref{fig:byci_skala_hw}) was deployed for several hours of flight time in the Bull Rock Cave located in the central Moravian Karst of the Czech Republic (see \autoref{fig:byci_skala_motivation} and the attached multimedia materials).

\begin{figure}[htb]
  \centering
  \subfloat[%
    Overview of the cave environment with the trajectories of all exploration missions (see \autoref{tab:byci_skala_exp}) performed within Bull Rock Cave.
    The figure shows deep cave missions (A, B), vertical flight (C), and the thorough exploration of a bounded area (D).
    ]{\hspace{-1.0em}\input{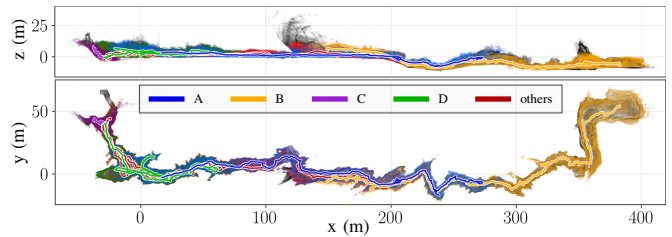}}
  \vspace{-0.5em}
  \subfloat[%
    Visual analysis on the mapping accuracy -- the distribution of mapping errors during all autonomous exploration tasks as summarized in \autoref{tab:byci_skala_exp}.
    The color bar legend represents the mapping error in meters using the point-to-point euclidean error metric.
    ]{\input{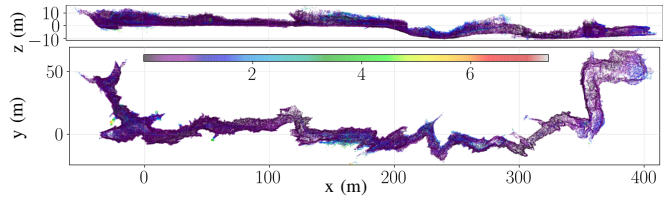}}
    \caption{%
      Full-coverage exploration of the Bull Rock Cave system (located in the central Moravian Karst, Czech Republic) with autonomous aerial explorers.
      Full resolution figure is available within the attached multimedia.}
    \label{fig:byci_skala_exp}
    \vspace{-1.5em}
\end{figure}

During multiple autonomous exploratory missions, a single explorer (see the hardware components of the robot in~\autoref{fig:byci_skala_hw}) was deployed to validate the proposed system in various exploratory scenarios.
The flight trajectories from all missions are visualized in~\autoref{fig:byci_skala_exp}a and the mission statistics and performance metrics of the mapping module are summarized in~\autoref{tab:byci_skala_exp}.
\rev{%
  A greedy frontier-navigation policy was employed such that the frontier closest to the lateral direction of flight (A, B), the highest frontier (C), and frontier with the largest ratio of unknown to free cells in a bounded area~(D) was selected as the next goal.}
With respect to these experiments in a harsh subterranean environment, we have
\begin{itemize}
  \item validated the performance of the system by flying in large cave domes, as well as in narrow corridors just \SI{70}{\centi\metre} wider than the dimensions of the robot,
  \item validated the real-time performance and robustness of the system in multiple autonomous horizontally-deep flights longer than \SI{470}{\metre} using just a single battery and reaching a maximal velocity up to \SI{2}{\metre\per\second},
  \item validated the ability to autonomously explore natural domes in terms of vertical depth,
  \item verified the ability to perform a full mission and return to an initial location with the obtained information,
  \item quantified the accuracy of the onboard-built maps with respect to a ground truth map of the environment, \rev{and}
  \item obtained feedback from speleologists in order to design the system following their requirements.
\end{itemize}

\rev{%
  The dense onboard-built maps ({\SI{20}{\centi\meter}} resolution) from all the experiments were merged (manual global registration with local ICP refinement) during post-processing to obtain the map of the environment $\mathcal{M}$.
The reference ground truth map $\mathcal{M}_{gt}$ was built by registering over~100 largely overlapping scans taken by a Leica BLK360 terrestrial 3D scanner.
The mapping accuracy over all the experiments reached mean $\mu = \SI{0.37}{\metre}$ and standard deviation $\sigma = \SI{0.46}{\metre}$ using the point-to-point Euclidean error metric between each point in $\mathcal{M}$ and the corresponding closest point in $\mathcal{M}_{gt}$}.
The distribution of the mapping errors throughout all flights is visualized in~\autoref{fig:byci_skala_exp}b.
As specified by the end-users, the decimeter-level mapping precision achieved over the course of these exceptionally fast and extensive flights is sufficient for the majority of speleological needs.

\sisetup{
  round-mode = places,
  round-precision = 0,
  table-parse-only,
  detect-weight=true,
  detect-inline-weight=text
}
\begin{table}[t]
  \scriptsize
  \centering
  \begin{tabular}{lrrrrr}
    \toprule
          & & & & \multicolumn{2}{c}{\scriptsize Mapping accuracy}\\ \cmidrule(r){5-6}
    Trial & \multirow{2}{*}[1.5em]{\thead[bc]{\scriptsize Flight\\\scriptsize time $\left(\si{\second}\right)$}} & \multirow{2}{*}[1.5em]{\thead[bc]{\scriptsize Trajectory\\\scriptsize length $\left(\si{\metre}\right)$}} & \multirow{2}{*}[1.5em]{\thead[bc]{\scriptsize Explored\\\scriptsize volume $\left(\si{\cubic\metre}\right)$}} & $\mu$ $\left(\si{\metre}\right)$ & $\sigma$ $\left(\si{\metre}\right)$\\\midrule
    A & \num{731.397362} & \num{475.513928} & \num{7463.256000} & \num[round-precision=2]{0.57} & \num[round-precision=2]{0.588} \\
    B & \num{934.700002} & \num{473.038235} & \num{11402.848000} & \num[round-precision=2]{0.529} & \num[round-precision=2]{0.5615}\\
    C & \num{358.500001} & \num{71.160049} & \num{550.864000} & \num[round-precision=2]{.233} & \num[round-precision=2]{.26}\\
    D & \num{749.298954} & \num{602.104045} & \num{3755.592000} & \num[round-precision=2]{.325} & \num[round-precision=2]{.383}\\
    \midrule
    E & \num{385.994784} & \num{233.249616} & \num{3054.568000} & \num[round-precision=2]{.387} & \num[round-precision=2]{.392}\\
    F & \num{633.499964} & \num{256.394032} & \num{2579.224000} & \num[round-precision=2]{.214} & \num[round-precision=2]{.218}\\
    G & \num{638.095865} & \num{261.296608} & \num{3650.432000} & \num[round-precision=2]{0.27} & \num[round-precision=2]{0.33}\\
    H & \num{297.099975} & \num{141.646622} & \num{1681.672000} & \num[round-precision=2]{.253} & \num[round-precision=2]{.406}\\
    I & \num{129.199385} & \num{121.243735} & \num{3325.784000} & \num[round-precision=2]{.194} & \num[round-precision=2]{.216}\\
    J & \num{424.600018} & \num{232.843902} & \num{4388.392000} & \num[round-precision=2]{.248} & \num[round-precision=2]{.291}\\
    \bottomrule
  \end{tabular}
  \caption{
    Quantitative evaluation on multiple autonomous exploratory missions within the Bull Rock Cave system.
    The flight trajectories and qualitative analysis of the mapping accuracy are shown in~\autoref{fig:byci_skala_exp}.
  }
  \label{tab:byci_skala_exp}
  \vspace{-2.0em}
\end{table}
\sisetup{
  round-precision=2,
}


\vspace{-0.2em}
\subsection{Virtual environment}

To validate the proposed methodology for multi-robot coordination using a local low-bandwidth communication network, a team of aerial robots was deployed for hundreds of hours of flight in a virtual environment using a virtual testbed developed for the DARPA Subterranean Challenge competition.
This state-of-the-art testbed consists of several large-scale cave environments containing dynamic obstacles and models of real-world interference, such as sensor discrepancies, communication schemes, and battery longevity.

\begin{figure}[t]
  \centering
  \input{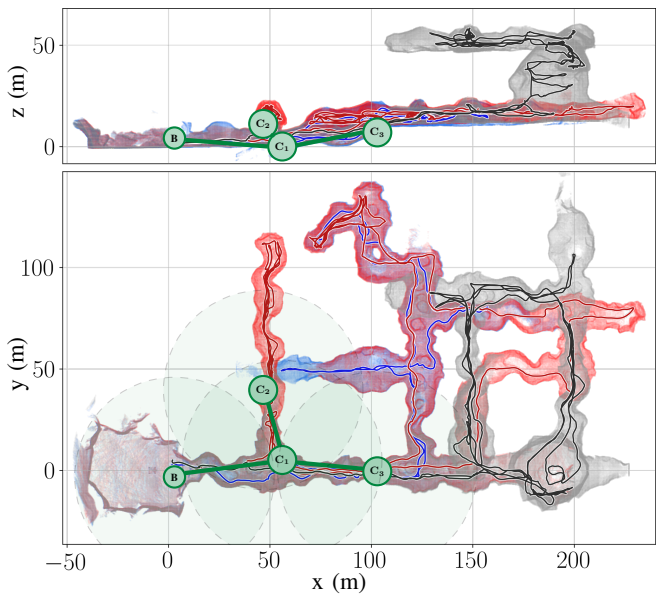}
  \vspace{-1.5em}
  \caption{
    Three autonomous explorers deployed in a virtual cave world within the DARPA simulation testbed.
    The robots finished their missions by building a communication tree with maximal edge length of $d_c = \SI{50}{\meter}$.
  }
  \label{fig:subt_run_18}
  \vspace{-0.8em}
\end{figure}

In contrast to real-world experiments, the virtual environment is larger and allows for the seamless verification of multi-robotic cooperation.
To demonstrate the performance of the proposed homing strategy, a selected example scenario of such an operation is presented in~\autoref{fig:subt_run_18}.
This experiment highlights the positive influence of the homing strategy in a search~\&~rescue scenario where the three explorers were able to exploit the increased flight time.
With a \SI{50}{\metre} communication range and \SI[round-precision=1]{1.2}{\metre\per\second} average velocity for each robot, the homing cooperation increased the available flight time for exploration by \SI{40}{\second} and \SI{80}{\second}, respectively.
\rev{%
  Moreover, the experiment shows the influence of multi-sensor mapping, which allowed the black robot to single-handedly explore the upper floor of the virtual environment.}
The final exploratory trajectories of the cooperating robots during the presented mission reached lengths of \SI{715}{\metre}, \SI{1349}{\metre}, and \SI{1405}{\meter}.

\rev{%
  The influence of the homing strategy on the time available for mere exploration is also quantitatively analyzed in~\mbox{\autoref{tab:homing}}.
  The results were averaged over six separate deployments, each with five cooperating robots.
  Identical mission parameters were set to all the robots for the baseline~{\cite{7139494}}, as well as for the proposed method.
  The data show an increasing trend in the available mission time for belated explorers for which the effective exploration phase is consequently prolonged during their entire operation time.}



\sisetup{
  round-mode = places,
  round-precision = 0,
  table-parse-only,
  detect-weight=true,
  detect-inline-weight=text
}
\begin{table}[t]
  \scriptsize
  \centering
  \begin{tabular}{lrrrrr}
    \toprule
    Robot                                            & 1$^{\text{st}}$ & 2$^{\text{nd}}$ & 3$^{\text{rd}}$ & 4$^{\text{th}}$ & 5$^{\text{th}}$\\
    Exploration time before homing (secs)            & \num{316.13} & \num{330.14} & \num{360.15} & \num{380.16} & \num{388.50}\\ 
    Exploration time increase $\left(\%\right)$      & \num[round-precision=1]{-1.5} & \num[round-precision=1]{2.84} & \num[round-precision=1]{12.196} & \num[round-precision=1]{18.43} & \num[round-precision=1]{21.03}\\ 
    \bottomrule
  \end{tabular}
  \caption{%
    Influence of the homing strategy on the flight time available for mere exploration.
    Comparison with a baseline time of {\SI{321}{\second}} (averaged over 10 flights) where a robot returned to base before its operation timed out.}
  \label{tab:homing}
  \vspace{-2.2em}
\end{table}




\section{Conclusion}
\label{sec:conclusion}


This letter presents a comprehensive study on the use of autonomous aerial explorers as an assisting technology for the exploration of natural cave environments.
This study also shares the experience acquired during the technology's development in close cooperation with a team of speleologists, cavers, and first responders.

The proposed self-sustaining system interconnects solutions for all crucial robotic tasks in order to enable full autonomy in complex unknown subterranean environments without access to GNSS.
Among others, this includes laser-data processing which copes with high humidity and dustiness within subterranean environments and robust path-planning for unknown dynamic environments to allow for flights in constrained cavities, as well as in open caverns of natural caves.
Moreover, a multi-robot cooperation is proposed for the efficient homing of a team of robots for applications where the possible information gain is superior to the costs of the robots, such as search~\&~rescue scenarios in cave systems.
The performance of the entire applicable system was validated in one of the most large-scale experimental analyses ever conducted, consisting of hours of flight time in Bull Rock Cave (Czech Republic, Moravian Karst) and in hundreds of hours in the state-of-the-art virtual testbed developed for the DARPA Subterranean Challenge.
This presented analysis of the entire system proves that it is a robust solution capable of reliable planning with sensoric field-of-view constraints and accurate mapping.
The accuracy of localization and mapping was evaluated with respect to a ground-truth map of the cave environment and reached mean precision below~\SI{40}{\centi\meter} in real-world conditions.
This performance has satisfied the requirements of speleologists and first responders.



\bibliographystyle{IEEEtran}
\bibliography{main}

\end{document}